\documentclass{article}
\usepackage{M-PSI}

\usepackage[english]{babel}
\usepackage{amsmath, amsthm, mathtools, amssymb, amsfonts}
\usepackage{csquotes}
\usepackage{bm}
\usepackage{graphicx}
\usepackage{xcolor}
\usepackage{booktabs}
\usepackage{multirow, makecell}
\usepackage{subcaption}
\usepackage{siunitx}
\usepackage{algorithm}
\usepackage{algpseudocode}
\usepackage{microtype}
\usepackage{float}
\usepackage{tikz}
\usepackage{textcomp}
\usepackage{url}

\usepackage[backend=biber,style=numeric-comp,sorting=none,natbib=true,maxbibnames=99,giveninits=true,doi=true,url=false,isbn=false,eprint=false]{biblatex}
\def\BibTeX{{\rm B\kern-.05em{\sc i\kern-.025em b}\kern-.08em
    T\kern-.1667em\lower.7ex\hbox{E}\kern-.125emX}}

\title{Anomalies in Multivariate Time Series Benchmarks\\Are Mostly Univariate}

\setsubtitle{Accepted at MiLeTS 2026 @ KDD -- Author Version}

\author{\href{https://marcpinet.fr/}{Marc Pinet\textsuperscript{1,2~\small{\ExternalLink}}}, \href{https://jcumin.github.io/}{Julien Cumin\textsuperscript{1~\small{\ExternalLink}}}, \href{https://www.researchgate.net/profile/Samuel-Berlemont}{Samuel Berlemont\textsuperscript{1~\small{\ExternalLink}}}, \href{https://research.vaufreydaz.org/}{Dominique Vaufreydaz\textsuperscript{2~\small{\ExternalLink}}}\vspace{0.1cm}\\
{$^1$ Orange Research, Meylan, France}\\
{$^2$ Univ. Grenoble Alpes, CNRS, Grenoble INP, LIG, 38000 Grenoble, France}\\
{Corresponding author: \href{mailto:marc.pinet@orange.com}{marc.pinet@orange.com}}
}

\hypersetup{
  hidelinks,
  urlcolor=red,
  pdftitle={Anomalies in Multivariate Time Series Benchmarks Are Mostly Univariate},
  pdfauthor={Marc Pinet, Julien Cumin, Samuel Berlemont, Dominique Vaufreydaz},
  pdfkeywords={Anomaly detection, Multivariate time series, Datasets, Evaluation methodology, Time series analysis, Benchmark reliability},
}

\begin{document}

\begin{abstract}
Many recent multivariate time series anomaly detection (MTSAD) models incorporate cross-channel modeling, under the implicit assumption that the structure of anomalies may be spread across multiple channels. We evaluate this assumption on eight widely used public benchmarks by introducing a per-segment diagnostic framework that flags, for each labeled anomaly, whether at least one channel deviates individually from its normal history, whether the cross-channel correlation structure changes, or both. The framework shows that no cross-channel rupture occurs without an accompanying univariate deviation across a range of reasonable thresholds. A complementary metric also reveals that on six of the eight benchmarks, at least half of the labeled anomaly segments deviate univariately on $89\%$ to $100\%$ of their timesteps, reaching $100\%$ on three of these datasets. To verify that our framework captures cross-channel structure when present, we construct synthetic data of phase-shifted sinusoidal channels with shared noise. Each anomalous segment is altered through one of two channel-wise corruptions that preserve the per-channel marginal distribution while breaking cross-channel structure, and our framework correctly characterizes these segments as cross-channel-only. On these data, channel-dependent (CD) models successfully exploit the cross-channel signal whereas channel-independent (CI) ones fail. The CI/CD comparison of a recent SOTA detector on real benchmarks further confirms that CD modeling brings no measurable gain. We conclude that current MTSAD benchmarks are unsuitable for validating cross-channel modeling capabilities, and we call for the development of more structurally diverse evaluation sets. The code for this study is publicly available\footnote{\url{https://github.com/marcpinet/mtsad-benchmarks-are-mostly-univariate}}.

\keywords{Anomaly detection \and Multivariate time series \and Datasets \and Evaluation methodology \and Time series analysis \and Benchmark reliability}
\end{abstract}

\section{Introduction}
\label{sec:introduction}
Many monitoring tasks rely on time series anomaly detection (TSAD)~\citep{paparrizosAdvancesTimeSeriesAnomaly2025}. In the multivariate setting (MTSAD), recent architectures have increasingly turned to channel-aware mechanisms: graph-based detectors~\citep{dengGraphNeuralNetworkBased2021, zhaoMultivariateTimeseriesAnomaly2020}, cross-channel modeling through learnable masked attention~\citep{wuCATCHChannelAwareMultivariate2025, zexerXCTFormerLeveragingCrossChannel2026}, and more. These models rely on the same implicit assumption: that public benchmarks contain anomalies that are more easily detected by modeling cross-channel dependencies. The closely related field of forecasting has reached the opposite consensus, where the channel-independent (CI) strategy popularized by PatchTST~\citep{nieTimeSeriesWorth2023} dominates through a capacity-robustness trade-off~\citep{hanCapacityRobustnessTradeoff2024}, and where the recent survey of~\citet{qiuComprehensiveSurveyDeep2026} formalizes the intermediate channel-partiality (CP) strategy. This shift has begun to propagate to TSAD, where several recent detectors adopt channel-independent designs~\citep{yangDCdetectorDualAttention2023, zhouKANADTimeSeries2025, liCrossADTimeSeries2025, shentuGeneralTimeSeries2025, chenCiTranGANChannelIndependentBasedAnomaly2025a} and report competitive or superior performance to channel-aware counterparts.

For each labeled anomalous segment in a multivariate benchmark, we ask: \emph{is there evidence of a univariate deviation (an individual channel deviating from its recent normal history), of a change in cross-channel dependence, both, or neither?} Our per-segment diagnostic (\autoref{sec:diagnostic-framework}) combines a univariate test with a cross-channel test based on the Pearson and Spearman correlations and the unbiased squared distance correlation~\citep{szekelyDistanceCorrelationTest2013}, with lagged extensions. Applied to eight standard benchmarks (\autoref{sec:empirical-evaluation}), the diagnostic finds no labeled segment in the strictly cross-channel category, on any benchmark, under any of the three measures, across a sensitivity sweep over both thresholds within their reasonable range. Every cross-channel rupture co-occurs with a univariate deviation, and, on six of the eight benchmarks, at least half of the labeled anomaly segments deviate univariately between $89\%$ and $100\%$ of their timesteps. A synthetic protocol (\autoref{sec:synthetic-validation}) confirms that the diagnostic identifies cross-channel ruptures when they exist. Moreover, channel-dependent models trained on these data recover the anomalous segments, whereas channel-independent variants perform at random (AUC-ROC $\approx 0.5$). On real benchmarks, our channel-dependent variant of the channel-independent state-of-the-art (SOTA) detector CrossAD~\citep{liCrossADTimeSeries2025} brings no measurable gain over the original (\autoref{sec:detection-real-benchmarks}).

Our contributions are: (i) a per-segment diagnostic framework that characterizes each labeled anomaly along two axes (univariate and cross-channel), with extension to lag and a sensitivity analysis over diagnostic thresholds; (ii) an empirical evaluation of eight standard benchmarks reporting the absence of strictly cross-channel anomaly segments, complemented by a univariate-ratio metric that measures how persistent the univariate signal is within each segment; (iii) a synthetic protocol validating the proposed diagnostic framework, on which a flattened linear autoencoder outperforms two recent SOTA detectors on anomalies that require cross-channel modeling; and (iv) a comparison on real benchmarks showing that the channel-dependent variant of CrossAD does not improve over its channel-independent counterpart, and in one case collapses entirely.

\section{Related Work}
\label{sec:related-work}
MTSAD models are classified by their treatment of cross-channel dependence into three strategies~\citep{qiuComprehensiveSurveyDeep2026}: channel-independent (CI), where each channel is processed independently without considering interactions among channels; channel-dependent (CD), which treats all channels as a unified entity assumed to be interdependent; and channel-partiality (CP), where each channel retains some independence while interacting selectively with a subset of others. In TSAD specifically, channel-aware (CD and CP) designs have proliferated, from graph-based detectors (GDN~\citep{dengGraphNeuralNetworkBased2021}, MTAD-GAT~\citep{zhaoMultivariateTimeseriesAnomaly2020}) to MLP-Mixer-based designs with explicit channel mixing, where a multi-layer perceptron (MLP) operates along the channel dimension to fuse cross-channel information (PatchAD~\citep{zhongPatchADLightweightPatchbased2025}, CCM-TAD~\citep{muradClusterAwareCausalMixer2025}) and cross-channel attention (XCTFormer~\citep{zexerXCTFormerLeveragingCrossChannel2026}, CATCH~\citep{wuCATCHChannelAwareMultivariate2025}). Hybrid strategies that combine a channel-independent prediction branch with a channel-mixing reconstruction branch (DBAD~\citep{sunDBADDualBranchTime2025}) further illustrate the variety of channel-strategy choices currently being explored. Beyond architecture, ChInf~\citep{wangChannelMattersEstimating2026} scores anomalies with per-channel influence functions, while remaining structurally channel-independent (further discussed in \autoref{sec:detection-real-benchmarks}).

A parallel line of work has questioned the reliability of public MTSAD benchmarks. \citet{wuCurrentTimeSeries2021} identified four classes of flaws in the Yahoo, Numenta, NASA, and SMD\footnote{The benchmark that~\citet{wuCurrentTimeSeries2021} call \enquote{OMNI} is the Server Machine Dataset introduced alongside OmniAnomaly by~\citet{suRobustAnomalyDetection2019a}.} benchmarks (triviality, unrealistic anomaly density, mislabeled ground truth, and run-to-failure bias) and issued five recommendations, including the abandonment of these datasets and the systematic visualization of data and algorithmic outputs. The authors target the quality of labeled segments at a per-exemplar and univariate level, whereas our evaluation is structural and explicitly multivariate: it asks, for each labeled segment, what kind of evidence (univariate or cross-channel) the data actually carries. These two works thus complement each other. \citet{wagnerTimeSeADBenchmarkingDeep2023} conducted the first audit dedicated to multivariate benchmarks and concluded that SWaT, WADI (water treatment and distribution), MSL, and SMAP (spacecraft telemetry) are unsuited for MTSAD evaluation due to distribution shift, positional bias, and the presence of a single continuous sensor channel alongside binary ones in MSL and SMAP. \citet{sarfrazPositionQuoVadis2024} showed that classical baselines such as Principal Component Analysis (PCA) reconstruction error and nearest-neighbor distance match or exceed SOTA deep learning models on standard MTSAD benchmarks, and further reduced trained deep TSAD models to linear distillations that retained their detection performance. \citet{liuElephantRoomReliable2024} curated TSB-AD across $40$ datasets and concluded that statistical methods and lightweight models often outperform SOTA deep learning detectors. Most recently, \citet{zhou2026mtsbench} introduced mTSBench and observed \enquote{high-magnitude anomalies in certain channels} for PSM, SMD, and SMAP.

Most directly relevant to our work, \citet{gargEvaluationAnomalyDetection2022a} claimed, based on author descriptions and visual inspection, that the standard MTSAD benchmarks \enquote{primarily contain temporal anomalies} and that \enquote{we did not find open datasets where cross-channel anomalies are known to be present}. They also observed that a univariate fully connected autoencoder beats SOTA multivariate algorithms across seven datasets. \citet{wenig2024} later reported the same pattern from a detector-comparison angle on fourteen benchmarks. Both works infer the structure of the benchmarks indirectly, through detector performance or visual inspection.

Across all these audits, the structure of the labeled anomalies themselves is inferred indirectly, through detector performance or qualitative case analysis. Our contribution is to characterize them directly through a per-segment diagnostic that operates on the data independently of any detector, and to validate it on a controlled synthetic protocol of cross-channel-only anomalies.

\section{Diagnostic Framework}
\label{sec:diagnostic-framework}
Our diagnostic framework operates on each anomalous segment and applies two tests: a univariate test that flags deviations from a recent normal history, and a cross-channel test that flags changes in the dependence structure between channels. \emph{The goal is to characterize the evidence available to a detector, rather than implementing an anomaly detector.} Unlike an anomaly detector, the diagnostic uses the ground-truth labels as input: it isolates each labeled segment, builds its normal history from the preceding unlabeled timesteps, and measures the evidence carried by the segment relative to that history.

\subsection{Definitions}
\label{subsec:definitions}

We denote by $\bm{X} \in \mathbb{R}^{T \times C}$ a multivariate time series of $C$ channels and $T$ timesteps indexed by $t \in \{1, \ldots, T\}$, with binary labels $\bm{y} \in \{0, 1\}^T$ where $y_t = 1$ marks an anomalous timestep.

An \emph{anomalous segment} $\mathcal{A} = [t_s, t_e]$ is any maximal contiguous run of timesteps with $y_t = 1$ in $\bm{y}$. Its length is denoted $|\mathcal{A}| = t_e - t_s + 1$. Each segment is analyzed independently (referred to as \emph{per-segment} throughout the paper).

For each anomalous segment $\mathcal{A}$, we define its \emph{normal history} $\mathcal{H}_{\mathcal{A}}$ as the $H$ closest normal timesteps preceding $t_s$. The history is constructed by walking backwards from $t_s$, skipping any anomalous timestep encountered along the way.

\subsection{Univariate Test}
\label{subsec:univariate-test}
For each anomalous segment $\mathcal{A}$, we estimate the per-channel normal mean $\mu_c^{\mathcal{H}_{\mathcal{A}}}$ and standard deviation $\sigma_c^{\mathcal{H}_{\mathcal{A}}}$ from $\mathcal{H}_{\mathcal{A}}$. We then compute, for every channel $c \in \{1, \ldots, C\}$, the maximum absolute z-score reached inside the segment:
\begin{equation}
z_c^{\mathcal{A}} = \max_{t \in \mathcal{A}} \frac{|X_{t,c} - \mu_c^{\mathcal{H}_{\mathcal{A}}}|}{\sigma_c^{\mathcal{H}_{\mathcal{A}}}}.
\label{eq:zscore}
\end{equation}

The segment is classified \emph{univariate} if there is a channel $ c \in \{1, \ldots, C\}$ such that $z_c^{\mathcal{A}} > \tau_z$. We use a default threshold $\tau_z = 3$. Indeed, Chebyshev's inequality guarantees that for any random variable $X$ with finite mean $\mu$ and finite standard deviation $\sigma$:
\begin{equation}
P(|X - \mu| > k\sigma) \leq \frac{1}{k^2}, \quad \forall k > 0.
\label{eq:chebyshev}
\end{equation}

With $\tau_z = 3$, the probability that a single timestep deviates from its normal mean by more than $3\sigma_c^{\mathcal{H}_{\mathcal{A}}}$ under non-anomalous stationary conditions is bounded by $1/9 \approx 11\%$, regardless of the distribution (this is only an upper bound, not the actual probability). The threshold is therefore not chosen to fit any benchmark, and we report a sensitivity analysis over $\tau_z \in \{2, 3, 4, 5, 7, 10\}$ in \autoref{subsec:classification-results}.

While this condition is necessary for a segment to be classified as univariate, it is not sufficient on its own: the existence of a single timestep exceeding $\tau_z$ in one channel says nothing about how pronounced the univariate signal is across $\mathcal{A}$. We therefore complement this binary test with a per-segment intensity measure in \autoref{subsec:univariate-ratio}.

We keep the z-score as the default test for its simplicity. When it is inapplicable, for instance on channels that are binary or nearly constant, we fall back to a Matrix Profile~\citep{yehMatrixProfileAll2016a} applied on each channel independently, which captures unusual motifs rather than unusual values.

The z-score is deliberately a minimal univariate test, capturing only deviations in level relative to the recent history. Anomalies that the z-score misses but a richer univariate test would catch only strengthen our claim: they are still univariate.

\subsection{Correlation Test}
\label{subsec:correlation-test}
The cross-channel test compares the dependence structure between channels inside an anomalous segment with the one observed in its normal history. For each segment $\mathcal{A}$ and its history $\mathcal{H}_{\mathcal{A}}$, we compute two correlation matrices $\bm{R}^{\mathcal{A}}, \bm{R}^{\mathcal{H}_{\mathcal{A}}} \in \mathbb{R}^{C \times C}$ and quantify their discrepancy\footnote{Channels with negligible variance ($\sigma < 10^{-4}$) in either $\mathcal{A}$ or $\mathcal{H}_{\mathcal{A}}$ are excluded from the cross-channel test, since their pairwise correlations are undefined.} by:
\begin{equation}
\Delta_{\rho}^{\max} = \max_{i \neq j} \left| R^{\mathcal{A}}_{ij} - R^{\mathcal{H}_{\mathcal{A}}}_{ij} \right|.
\label{eq:delta-rho}
\end{equation}

The segment is classified \emph{cross-channel} if $\Delta_{\rho}^{\max} > \tau_{\rho}$ for at least one of the three following correlation methods:
\begin{itemize}
    \item the \textbf{Pearson} correlation and its coefficient $\rho_{ij} \in [-1, 1]$, measuring linear dependence,
    \item the \textbf{Spearman} rank correlation and its coefficient $\rho_{ij} \in [-1, 1]$, measuring monotone dependence,
    \item the \textbf{Distance correlation}~\citep{szekelyMeasuringTestingDependence2007a}, in its unbiased squared form~\citep{szekelyDistanceCorrelationTest2013} $\rho_{ij} = \mathrm{dCor}^2_U \in [-1, 1]$ (negative values are bias-correction noise, clipped to zero~\citep{monroyCastillo2025}), which captures any form of statistical dependence and is harder to trigger than Pearson/Spearman at high $\tau_\rho$.
\end{itemize}
The threshold $\tau_\rho \in [0, 2]$ controls the sensitivity of the cross-channel test. We use a permissive default threshold $\tau_\rho = 0.1$ and report a sensitivity analysis in \autoref{subsec:classification-results}. We use three measures rather than dCor alone because dCor is computed at lag zero only, while Pearson and Spearman allow an efficient lagged extension that captures lead-lag dependencies.

\paragraph{Extension to lag} Channel pairs in real-world systems often exhibit time-shifted relationships~\citep{zhaoRethinkingChannelDependence2024}. To capture such dependencies, we extend Pearson and Spearman to lagged correlations: $\hat{\rho}_{ij}(\ell)$ measures, on the training set, the correlation between channel $i$ at time $t$ and channel $j$ at time $t + \ell$, for $\ell \in \{-L_{\max}, \ldots, L_{\max}\}$.

Let $\mathcal{L} = \{-L_{\max}, \ldots, L_{\max}\} \setminus \{0\}$ denote the set of non-zero lags. A pair $(i, j)$ is recorded as \emph{lagged-dominant} if $\exists\, \ell^{\star}_{ij} \in \mathcal{L}$ such that $|\hat{\rho}_{ij}(\ell^{\star}_{ij})| > |\hat{\rho}_{ij}(0)|$. When computing $\bm{R}^{\mathcal{A}}$ and $\bm{R}^{\mathcal{H}_{\mathcal{A}}}$ in Equation~\eqref{eq:delta-rho}, the entry $(i, j)$ for any lagged-dominant pair uses the lagged correlation at $\ell^{\star}_{ij}$ rather than at lag zero.

Distance correlation is intentionally not extended to lags. Computing $\mathrm{dCor}^2_U$ at every lag for every channel pair was prohibitive even on the smallest benchmark in both channel count and series length.

\subsection{Classification}
\label{subsec:classification}
Combining both tests yields four per-segment labels: \textbf{UNIVARIATE}, \textbf{CROSS-CHANNEL}, \textbf{BOTH}, or \textbf{UNDETECTED}. Segments shorter than $L_{\min}$ skip the cross-channel test (\autoref{subsec:correlation-test}) and fall back to UNIVARIATE or UNDETECTED. A CROSS-CHANNEL segment that fails the univariate test at every timestep would thus require channel-aware modeling.

\subsection{Univariate Ratio}
\label{subsec:univariate-ratio}
The classification of \autoref{subsec:classification} flags whether at least one timestep in a segment exhibits a univariate deviation, but it does not measure how persistent this deviation is across the segment (in the same way that a single positive prediction inflates point-adjusted metrics~\citep{kimRigorousEvaluationTimeseries2022}). To quantify this persistence, we define the per-segment \emph{univariate ratio}:
\begin{equation}
u^{\mathcal{A}} = \frac{1}{|\mathcal{A}|} \sum_{t \in \mathcal{A}} \mathbf{1}\!\left\{ \max_{c \in \{1, \ldots, C\}} \frac{|X_{t,c} - \mu_c^{\mathcal{H}_{\mathcal{A}}}|}{\sigma_c^{\mathcal{H}_{\mathcal{A}}}} > \tau_z \right\}.
\label{eq:uni-ratio}
\end{equation}
The ratio $u^{\mathcal{A}} \in [0, 1]$ is the fraction of timesteps within the segment where at least one channel exceeds the z-score threshold. A value of $u^{\mathcal{A}} = 1$ means that every timestep of the segment exhibits a univariate deviation on at least one channel. A value close to zero means that the segment passes the binary univariate test due to a single or a few isolated timesteps. We report $u^{\mathcal{A}}$ both globally (over all long segments) and restricted to segments classified as BOTH, which directly addresses whether segments flagged by both tests retain a persistent univariate signal or a sparse one. Sensitivity to $\tau_z$ is reported in \autoref{subsec:univariate-ratio-results}.

\section{Empirical Evaluation}
\label{sec:empirical-evaluation}

\subsection{Datasets}
\label{subsec:datasets}
We evaluate eight benchmarks used in the vast majority of recent MTSAD evaluations~\citep{paparrizosAdvancesTimeSeriesAnomaly2025, sarfrazPositionQuoVadis2024, wagnerTimeSeADBenchmarkingDeep2023, schmidlAnomalyDetectionTime2022}: GECCO~\citep{moritzGECCOIndustrialChallenge2018}, NASA (MSL and SMAP)~\citep{hundmanDetectingSpacecraftAnomalies2018}, PSM~\citep{abdulaalPracticalApproachAsynchronous2021}, SMD~\citep{suRobustAnomalyDetection2019a}, SWaT~\citep{gohSWaTDataset2017}, SWAN-SF~\citep{angrykSWANSFSpaceWeather2020}, and WADI~\citep{ahmedWADIWaterDistribution2017}. They span water treatment and distribution (GECCO, SWaT, WADI), spacecraft telemetry (MSL, SMAP), server metrics (PSM, SMD), and solar weather (SWAN-SF). Their main characteristics are summarized in \autoref{tab:datasets}.

\begin{table}[t]
    \centering
    \small
    \caption{Summary of the eight MTSAD benchmarks evaluated. $C$ is the number of channels. The anomaly ratio is computed on the test split.}
    \begin{tabular}{lrrrr}
        \toprule
        Dataset & $C$ & Train size & Test size & Anomaly ratio (\%) \\
        \midrule
        GECCO    & 9  & 69\,260  & 69\,261  & 1.05 \\
        MSL      & 55 & 58\,317  & 73\,729  & 10.53 \\
        SMAP     & 25 & 140\,825 & 444\,035 & 12.87 \\
        PSM      & 25 & 132\,481 & 87\,841  & 27.76 \\
        SMD      & 38 & 708\,405 & 708\,420 & 4.16 \\
        SWAN-SF  & 38 & 60\,000  & 60\,000  & 32.60 \\
        SWaT     & 31 & 472\,459 & 472\,460 & 11.56 \\
        WADI     & 79 & 784\,571 & 172\,803 & 5.77 \\
        \bottomrule
    \end{tabular}
    \label{tab:datasets}
\end{table}

We adopt a uniform preprocessing pipeline: for each dataset, we fit a per-channel \texttt{StandardScaler} on the training set and apply the resulting transform to both train and test splits. SMD is split into 28 separate machines and each machine is processed independently, with its own scaler, as recommended by~\citet{suRobustAnomalyDetection2019a}. The GECCO training set contains $1.44\%$ of labeled anomalies, which we filter out before the preprocessing. The original train/test splits are preserved.

\subsection{Setup}
\label{subsec:setup}
We apply the diagnostic of \autoref{sec:diagnostic-framework} with the following default hyperparameters: history length $H = 300$, univariate threshold $\tau_z = 3$, correlation threshold $\tau_\rho = 0.1$, maximum lag $L_{\max} = 192$, and minimum segment length $L_{\min} = 10$. We verified that the conclusions of this section are stable for $H \in \{100, 300, 1000\}$. The permissive correlation threshold $\tau_\rho$ deliberately makes the cross-channel test easy to trigger, which strengthens the central claim of this paper: the lack of segments classified as cross-channel without a univariate deviation. The maximum lag $L_{\max}$ matches the upper end of sliding window sizes used in recent MTSAD models~\citep{liCrossADTimeSeries2025, wuCATCHChannelAwareMultivariate2025}. Sensitivity to $\tau_z$ and $\tau_\rho$ is reported in \autoref{subsec:classification-results} and \autoref{subsec:univariate-ratio-results}.

\subsection{Classification Results}
\label{subsec:classification-results}

\begin{table}[t]
\centering
\small
\caption{Per-segment classification on the eight benchmarks, for each of the three correlation methods (P = Pearson, S = Spearman, D = $\mathrm{dCor}^2_U$). The chosen thresholds are $\tau_z=3, \tau_\rho=0.1$. Long segments are those with $|\mathcal{A}| \geq L_{\min} = 10$. The two rightmost columns count short segments labeled UNIVARIATE or UNDETECTED. The CROSS-CHANNEL category is empty everywhere, on every method, and is therefore not shown.}
\label{tab:classification-results}
\begin{tabular}{l rrr rrr r r}
\toprule
& \multicolumn{3}{c}{UNIVARIATE} & \multicolumn{3}{c}{BOTH} & \multicolumn{2}{c}{Short} \\
\cmidrule(lr){2-4} \cmidrule(lr){5-7} \cmidrule(lr){8-9}
Dataset & P & S & D & P & S & D & UNI & UND \\
\midrule
GECCO   & 0  & 0  & 0  & 22  & 22  & 22  & 0    & 0    \\
MSL     & 4  & 3  & 31 & 30  & 31  & 3   & 0    & 0    \\
SMAP    & 15 & 15 & 16 & 41  & 41  & 40  & 0    & 0    \\
PSM     & 0  & 0  & 0  & 34  & 34  & 34  & 33   & 4    \\
SMD     & 0  & 0  & 0  & 176 & 176 & 176 & 148  & 1    \\
SWAN-SF & 0  & 0  & 0  & 2   & 2   & 2   & 5079 & 1056 \\
SWaT    & 0  & 0  & 0  & 35  & 35  & 35  & 0    & 0    \\
WADI    & 0  & 0  & 0  & 14  & 14  & 14  & 0    & 0    \\
\bottomrule
\end{tabular}
\end{table}

\autoref{tab:classification-results} reports the per-segment classification on each benchmark, with each of the three correlation methods. Out of $373$ long segments aggregated across all datasets and methods, the CROSS-CHANNEL category is \emph{empty}: every detected cross-channel rupture co-occurs with a univariate deviation. The BOTH category dominates almost everywhere. This is consistent with our permissive correlation threshold $\tau_\rho = 0.1$.

\begin{figure*}[t]
    \centering
    \includegraphics[width=\textwidth]{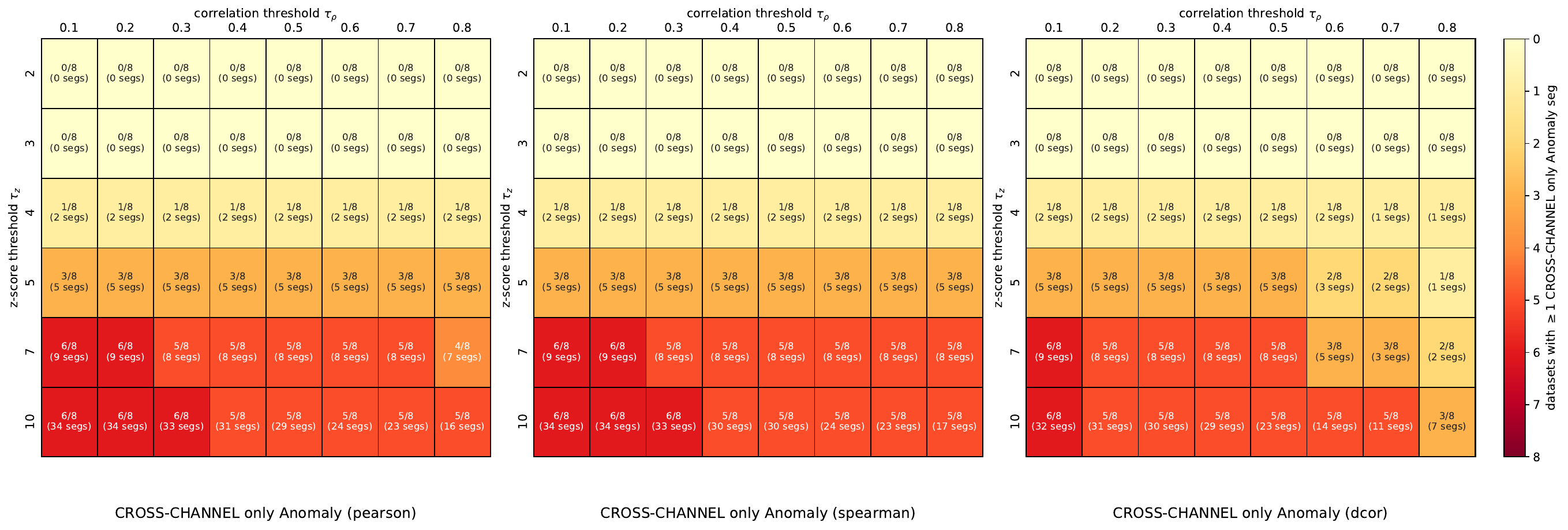}
    \caption{Sensitivity of the CROSS-CHANNEL count to both thresholds $\tau_z$ and $\tau_{\rho}$, aggregated across the eight benchmarks for each of the three correlation methods. Each cell reports the number of datasets containing at least one CROSS-CHANNEL segment, followed by the total count of such segments in parentheses.}
    \label{fig:heatmap-classif}
\end{figure*}

\autoref{fig:heatmap-classif} reports the CROSS-CHANNEL count for each correlation method across different $(\tau_z, \tau_\rho)$, aggregated across the eight benchmarks. At $\tau_z \in \{2, 3\}$, zero CROSS-CHANNEL segments are detected at any $\tau_\rho \in [0.1, 0.8]$ on any benchmark. At $\tau_z = 4$, two CROSS-CHANNEL segments appear on a single dataset. It is only at $\tau_z \geq 7$, where Chebyshev's bound drops below $1/49 \approx 2\%$, that CROSS-CHANNEL segments appear on a majority of benchmarks.

Even at extreme thresholds, the CROSS-CHANNEL category remains small: across the entire $(\tau_z, \tau_\rho)$ sweep, the maximum number of CROSS-CHANNEL segments observed in a single configuration is $34$ out of the $373$ long segments aggregated across all benchmarks, that is less than $\approx{9}\%$.

\begin{figure}[t]
    \centering
    \includegraphics[width=\linewidth]{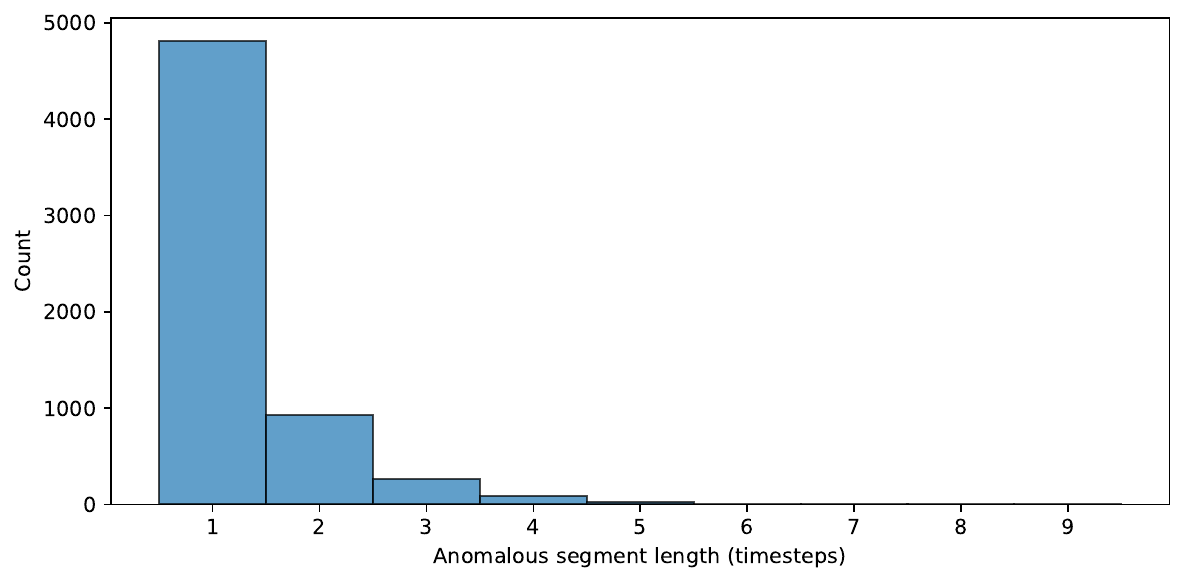}
    \caption{Distribution of segment lengths on SWAN-SF. All labeled anomalies are point-wise (length $1$) or near-point-wise, with only two segments exceeding $L_{\min} = 10$.}
    \label{fig:swan-short-distrib}
\end{figure}

\paragraph{SWAN-SF} Out of its $6\,137$ labeled segments, only $2$ exceed $L_{\min} = 10$. The remaining $6\,135$ segments are too short for the cross-channel test and thus are only classified by the univariate test (\autoref{subsec:correlation-test}); their length distribution is reported in \autoref{fig:swan-short-distrib}. Among them, $5\,079$ ($83\%$) are classified UNIVARIATE, $1\,056$ are UNDETECTED, and $2$ lack sufficient context (at the beginning of the test set, where fewer than $H = 300$ normal points precede the segment). The $1\,056$ UNDETECTED segments do not threaten the main claim of \autoref{sec:empirical-evaluation}, being too short for the cross-channel test.

\paragraph{MSL and SMAP} The NASA benchmarks differ from the others. All channels in MSL and SMAP are binary (with the exception of one channel each), which limits the discriminative power of the z-score test. We therefore fall back to the Matrix Profile (see \autoref{subsec:univariate-test}) to confirm that these segments, especially the BOTH ones, remain detectable univariately. On MSL, it detects $27$ out of $31$ BOTH segments (union over the three correlation methods), while on SMAP, it identifies $37$ out of $43$ segments. The remaining segments are visually indistinguishable from their normal history, consistent with the labeling concerns raised by~\citet{wuCurrentTimeSeries2021}.

\subsection{Univariate Ratio Results}
\label{subsec:univariate-ratio-results}

\begin{figure}[t]
    \centering
    \includegraphics[width=\linewidth]{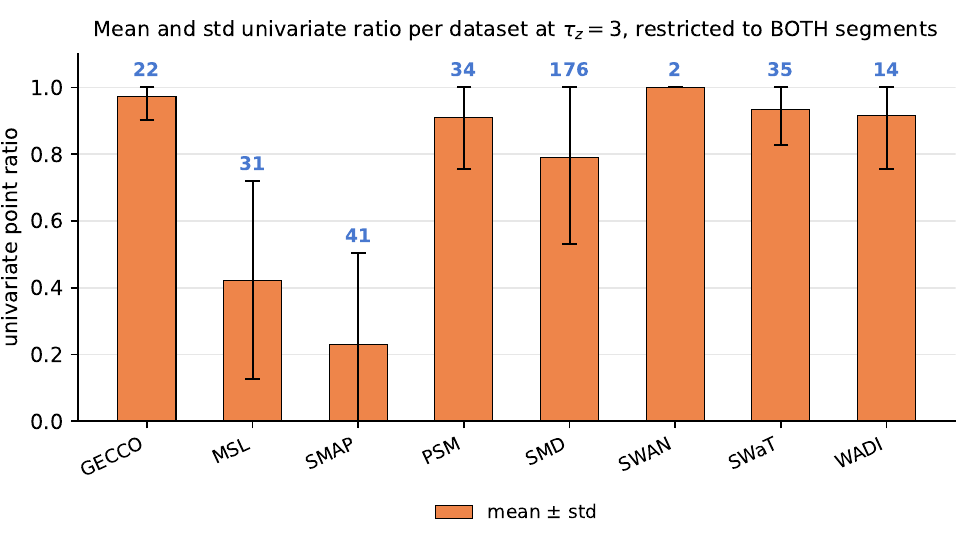}
    \caption{Mean univariate ratio (with standard deviation) per dataset at $\tau_z = 3$, restricted to BOTH segments. The number above each bar is the segment count.}
    \label{fig:univariate-ratio-per-dataset}
\end{figure}

\autoref{fig:univariate-ratio-per-dataset} reports the mean (with standard deviation) univariate ratio per dataset at $\tau_z = 3$, restricted to BOTH segments. We focus on this subset because it directly addresses our central question, i.e., whether segments flagged as both univariate and cross-channel have a dense univariate signal or not.

Six of the eight benchmarks have a mean univariate ratio between $0.79$ and $1.00$. Labeled anomalies on these benchmarks therefore not only deviate univariately somewhere within their anomalous segments, but deviate univariately on most of their anomalous timesteps. MSL and SMAP exhibit lower means ($0.42$ and $0.23$ respectively) and are discussed below. Note that SWAN-SF has only two long segments (its $6\,137$ short segments were analyzed in \autoref{subsec:classification-results}).

\begin{figure}[t]
    \centering
    \includegraphics[width=\linewidth]{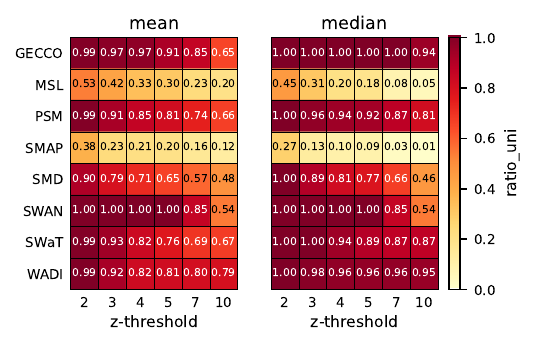}
    \caption{Sensitivity of the univariate ratio to $\tau_z$ across the eight benchmarks, restricted to BOTH segments. Mean (left) and median (right).}
    \label{fig:uni-ratio-sweep}
\end{figure}

\autoref{fig:uni-ratio-sweep} reports the univariate ratio on BOTH segments with $\tau_z \in \{2, 3, 4, 5, 7, 10\}$. Raising $\tau_z$ from $3$ to $5$, which tightens Chebyshev's bound from $1/9 \approx 11\%$ to $1/25 = 4\%$, leaves the median ratio above $0.89$ on five of the eight benchmarks (GECCO, PSM, SWAN, SWaT, and WADI) and at $0.77$ on SMD. The same pattern holds across the entire sweep: even at $\tau_z = 10$ where Chebyshev's bound drops to $1\%$, the median ratio stays at $0.94$, $0.81$, $0.87$, and $0.95$ on GECCO, PSM, SWaT, and WADI respectively.

Both NASA benchmarks exhibit low univariate ratios across all thresholds, with SMAP medians dropping to $0.13$ already at $\tau_z = 3$. As explained in \autoref{subsec:classification-results}, the z-score is not suited to capture how strongly they deviate univariately. However, the complementary Matrix Profile evaluation reported in the aforementioned section already confirms that the BOTH segments remain detectable univariately.

The diagnostic of \autoref{sec:empirical-evaluation} reports an empty CROSS-CHANNEL column on every benchmark, but this raises an immediate concern: would the diagnostic actually flag a CROSS-CHANNEL segment if one existed? We address this with a controlled synthetic protocol in which anomalies are strictly cross-channel, then leverage the same protocol to compare CI, CD, and CP detectors in a setting where cross-channel modeling is required.

\section{Validation Protocol using Synthetic Data}
\label{sec:synthetic-validation}
The evaluation is complemented with a controlled synthetic protocol in which every anomaly is explicitly constructed to be strictly cross-channel.

\subsection{Construction of Synthetic Data}
\label{subsec:construction}
We generate a synthetic dataset of $C = 9$ phase-shifted sinusoidal channels with shared additive noise. For each timestep $t$ and channel $c$:
\begin{equation}
X_{t,c} = \sin\!\left(\frac{2\pi t}{P} + \frac{2\pi c}{C}\right) + \varepsilon_t,
\quad \varepsilon_t \stackrel{\mathrm{iid}}{\sim} \mathcal{N}(0, \sigma_\varepsilon^2),
\label{eq:synthetic-clean}
\end{equation}

with period $P = 50$ and noise standard deviation $\sigma_\varepsilon = 0.3$. $\varepsilon_t$ does not depend on $c$: the noise is shared across channels at each timestep. This induces a strong correlation between every pair of channels in the clean signal. The training set has $T_{\mathrm{train}} = 20\,000$ anomaly-free timesteps. The test set has $T_{\mathrm{test}} = 10\,000$ timesteps and contains four anomalous segments of lengths $400$, $300$, $500$, and $300$, distributed at fixed timesteps.

For each anomalous segment $\mathcal{A}$, a subset $\mathcal{S}_\mathcal{A} \subset \{1, \ldots, C\}$ of channels is selected to be corrupted, uniformly at random. The size $|\mathcal{S}_\mathcal{A}|$ is itself drawn uniformly in $\{1, \ldots, C-1\}$. Both extremes are excluded because corrupting zero channels would leave the segment unaltered, while corrupting all channels would apply the same transformation to every channel and therefore leave the cross-channel correlation structure unchanged. We consider two corruption methods:
\begin{itemize}
    \item \textbf{NOISEFLIP.} For each $c \in \mathcal{S}_\mathcal{A}$ and $t \in \mathcal{A}$, set $X_{t,c} \leftarrow X_{t,c} - 2\varepsilon_t$, replacing the shared noise on channel $c$ by its opposite $-\varepsilon_t$. By symmetry of the Gaussian, $\varepsilon_t$ and $-\varepsilon_t$ are identically distributed, leaving the per-channel marginal unchanged while making the noise on $c$ anti-correlated with that of unaffected channels.
    \item \textbf{NPROLL.} For each $c \in \mathcal{S}_\mathcal{A}$, the values of channel $c$ within $\mathcal{A}$ are circularly shifted by a random offset drawn independently and uniformly in $\{1, \ldots, |\mathcal{A}|-1\}$. The segment's per-channel marginal is preserved, but both the phase of the sinusoidal component and the shared noise are temporally misaligned with unaffected channels.
\end{itemize}

\begin{figure}[t]
    \centering
    \begin{subfigure}{0.48\linewidth}
        \includegraphics[width=\linewidth]{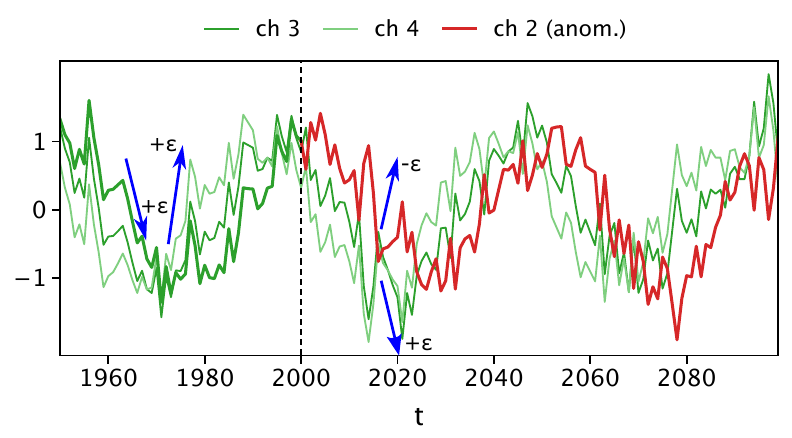}
        \caption{NOISEFLIP.}
        \label{fig:noiseflip}
    \end{subfigure}
    \hfill
    \begin{subfigure}{0.48\linewidth}
        \includegraphics[width=\linewidth]{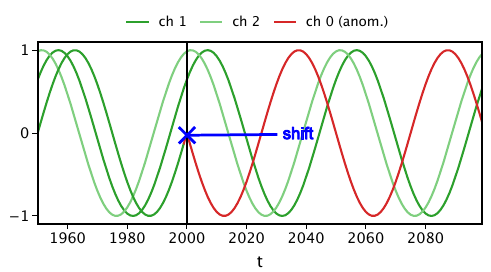}
        \caption{NPROLL.}
        \label{fig:nproll}
    \end{subfigure}
    \caption{Synthetic protocols that generate cross-channel-only anomalies. NPROLL has its noise removed for clearer view.}
    \label{fig:synthetic-injections}
\end{figure}

Both injection methods preserve every per-channel marginal distribution while breaking cross-channel structure. No univariate test operating on a single channel can flag the resulting segments, since each channel, whether in $\mathcal{S}_\mathcal{A}$ or not, exhibits the same per-segment marginal as in $\mathcal{H}_{\mathcal{A}}$. The only evidence available is the alteration of cross-channel dependence. These two protocols therefore provide a controlled cross-channel-only setting, in which our diagnostic should classify every anomalous segment as CROSS-CHANNEL.

\subsection{Diagnostic Validation}
\label{subsec:diagnostic-validation}

For each of the two injection methods of \autoref{subsec:construction}, we generate $1\,000$ independent test sets. Each test set contains $4$ anomalous segments, for a total of $4\,000$ segments per injection method. Every segment is then classified by the diagnostic of \autoref{sec:diagnostic-framework}, under each of the three correlation methods, with default thresholds $\tau_z = 3$ and $\tau_\rho = 0.1$. We expect the classification to be CROSS-CHANNEL on every segment, since both methods break cross-channel structure while preserving every per-channel marginal.

\begin{table}[t]
\centering
\small
\caption{Classification counts on the synthetic protocol, aggregated over $4,000$ segments per cell ($1,000$ seeds $\times$ $4$ segments). P, S, D denote Pearson, Spearman, and $\mathrm{dCor}^2_U$. UNIVARIATE and UNDETECTED counts are zero in every cell and therefore omitted.}
\label{tab:synthetic-classification}
\resizebox{\columnwidth}{!}{%
\begin{tabular}{l rrr rrr}
\toprule
& \multicolumn{3}{c}{NOISEFLIP} & \multicolumn{3}{c}{NPROLL} \\
\cmidrule(lr){2-4} \cmidrule(lr){5-7}
Category & P & S & D & P & S & D \\
\midrule
CROSS-CHANNEL & 3\,967 & 3\,967 & 3\,967 & 3\,970 & 3\,970 & 3\,970 \\
BOTH          & 33     & 33     & 33     & 30     & 30     & 30 \\
\bottomrule
\end{tabular}%
}
\end{table}

Across the $24,000$ classifications, $99.2\%$ of segments are classified as CROSS-CHANNEL and $0.8\%$ as BOTH, never UNIVARIATE nor UNDETECTED. NOISEFLIP produces moderate ruptures ($\Delta_\rho^{\max} \in [0.33, 0.40]$) while NPROLL produces stronger ones ($\approx 0.62$ for dCor, $\approx 1.5$ for Pearson/Spearman), all correctly classified. The average $z_{\max} \approx 2.52$ matches the value expected on a non-corrupted segment under the sinusoidal generative process, confirming marginal preservation. The residual BOTH segments result from rare timesteps where the shared Gaussian noise $\varepsilon_t$ happens to exceed $\tau_z$ standard deviation by chance. The near-total absence of CROSS-CHANNEL segments observed in \autoref{subsec:classification-results} on the eight real benchmarks therefore reflects the actual composition of the labeled anomalies.

\subsection{Detection on Synthetic Anomalies}
\label{subsec:detection-synthetic}
We use the synthetic protocol to compare channel-dependent (CD), channel-independent (CI), and channel-partiality architectures. Detectors operate on fixed-length windows: we denote a window by $\bm{W} \in \mathbb{R}^{L \times C}$ and its $c$-th channel by $\bm{w}_c \in \mathbb{R}^{L}$, where $L$ is the window length. We evaluate four reconstruction-based detectors, two of which are recent SOTA: \textbf{LinearAE} (flattened linear autoencoder), \textbf{AE} (two-layer MLP with GELU), \textbf{CrossAD}~\citep{liCrossADTimeSeries2025} (originally CI, for which we also implemented a CD variant), and \textbf{CATCH}~\citep{wuCATCHChannelAwareMultivariate2025} (CP). We pick CrossAD because its native CI design allows minimal modifications to obtain a CD variant (full implementation in the released code). For LinearAE, AE, and CrossAD, the CD variant flattens the window across channels before encoding while the CI variant shares the encoder across channels.

For each (model, corruption) pair, we sweep latent dimensions $d_{\mathrm{model}} \in \{16, 32, 64, 128\}$ and window sizes $L \in \{16, 32, 64, 128\}$ ($16$ configurations) and repeat each on three random seeds. CrossAD and CATCH use their authors' default training configurations, while LinearAE and AE train for $30$ epochs. All models train on the same $20\,000$ anomaly-free timesteps, with stride $1$ windows for training and non-overlapping windows for evaluation, so every test timestep receives exactly one reconstruction error. Experiments ran on an NVIDIA A100 (80 GB).

We report only AUC-ROC and AUC-PR. The range-based and volume-based extensions of~\citet{paparrizosVolumeSurfaceNew2022} were designed to handle range anomalies and label imprecision near anomaly boundaries; both are controlled here.

\begin{table}[t]
\centering
\small
\caption{Detection results on the synthetic protocol, mean $\pm$ standard deviation over $48$ runs on $3$ seeds.}
\label{tab:detection-results}
\resizebox{\columnwidth}{!}{%
\begin{tabular}{ll cc cc}
\toprule
& & \multicolumn{2}{c}{NPROLL} & \multicolumn{2}{c}{NOISEFLIP} \\
\cmidrule(lr){3-4} \cmidrule(lr){5-6}
Model & Var. & AUC-ROC & AUC-PR & AUC-ROC & AUC-PR \\
\midrule
\multicolumn{6}{l}{\emph{Channel-dependent}} \\
LinearAE & CD  & $0.97 \pm 0.05$ & $0.89 \pm 0.16$ & $0.84 \pm 0.17$ & $0.64 \pm 0.35$ \\
AE       & CD  & $0.97 \pm 0.05$ & $0.88 \pm 0.16$ & $0.81 \pm 0.18$ & $0.59 \pm 0.33$ \\
CrossAD  & CD  & $0.85 \pm 0.09$ & $0.61 \pm 0.19$ & $0.55 \pm 0.04$ & $0.21 \pm 0.02$ \\
\midrule
\multicolumn{6}{l}{\emph{Channel-partiality}} \\
CATCH    & CP & $0.77 \pm 0.11$ & $0.37 \pm 0.20$ & $0.66 \pm 0.11$ & $0.27 \pm 0.12$ \\
\midrule
\multicolumn{6}{l}{\emph{Channel-independent}} \\
LinearAE & CI  & $0.57 \pm 0.06$ & $0.17 \pm 0.02$ & $0.50 \pm 0.01$ & $0.15 \pm 0.01$ \\
AE       & CI  & $0.60 \pm 0.06$ & $0.18 \pm 0.03$ & $0.52 \pm 0.02$ & $0.16 \pm 0.01$ \\
CrossAD  & CI  & $0.53 \pm 0.02$ & $0.16 \pm 0.01$ & $0.47 \pm 0.01$ & $0.15 \pm 0.01$ \\
\bottomrule
\end{tabular}%
}
\end{table}

All three CI configurations and both protocols yield AUC-ROC in $[0.47, 0.60]$ and AUC-PR in $[0.15, 0.18]$ with little variance (see \autoref{tab:detection-results}). As expected from their respective inductive biases, CI models cannot detect these anomalies regardless of capacity. Note that we report these results not as a contribution but as a sanity check that our protocol indeed isolates cross-channel structure.

LinearAE-CD and AE-CD reach mean AUC-PR of $0.89$ and $0.88$ on NPROLL, with best configurations reaching $0.99$, well above CrossAD-CD ($0.61$) and CATCH ($0.37$). The same conclusions hold for NOISEFLIP. A flattened linear autoencoder across channels is enough to reach high performance. CATCH reaches performance in between CI and CD approaches. The large standard deviation of LinearAE (CD) and AE (CD) on NOISEFLIP reflects a sharp diagonal transition in the $(d, L)$ grid: AUC-ROC plateaus near $1.0$ when $d_{\mathrm{model}} \geq L$ and collapses below $0.7$ otherwise, mixing near-perfect and near-random runs across the $16$ configurations.

\begin{figure}[t]
    \centering
    \includegraphics[width=\linewidth]{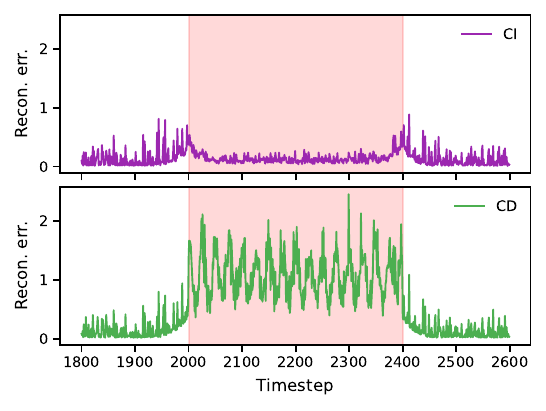}
    \caption{Reconstruction losses from LinearAE (CI and CD) on an NPROLL anomaly segment. CI captures the boundaries but not the whole segment whereas CD fully captures it. Latent dimension is 16 and window size is 64.}
    \label{fig:ci-cd-recon-loss}
\end{figure}

NPROLL is easier to detect than NOISEFLIP as it introduces sharp discontinuities at the segment boundaries, which a reconstruction-based detector easily flags. NOISEFLIP leaves both the sinusoidal signal and the segment endpoints intact, with only the small shared-noise component flipped on affected channels. We can observe this phenomenon in the reconstruction losses (see \autoref{fig:ci-cd-recon-loss}).

A flattened linear autoencoder outperforming two SOTA models, including one specifically designed for cross-channel modeling, raises a concern: if complex channel-aware mechanisms do not help even where cross-channel modeling is necessary, what do they contribute on benchmarks where every labeled anomaly is already univariately detectable (\autoref{sec:empirical-evaluation})?

\section{Detection on Real Benchmarks}
\label{sec:detection-real-benchmarks}
From the evaluations reported in \autoref{sec:empirical-evaluation}, we could hypothesize that CI and CD variants of the same model should behave similarly on real benchmarks, since no labeled anomaly requires cross-channel modeling for detection. We test this prediction directly by training CrossAD~\citep{liCrossADTimeSeries2025} in both its original CI form and our CD variant on the same datasets as in the original CrossAD evaluation (seven of the eight benchmarks of \autoref{sec:empirical-evaluation}, WADI excluded). Following the CrossAD protocol, we only retain the first channel of MSL and SMAP. We train both variants with the default configuration of CrossAD for each dataset, using their codebase directly (and therefore do not duplicate it in our own repository). \autoref{tab:crossad-results} reports the results on VUS metrics~\citep{paparrizosVolumeSurfaceNew2022} as recommended by~\citet{liuElephantRoomReliable2024} for real-world settings where anomaly boundaries are often uncertain.

\begin{table}[t]
\centering
\caption{CrossAD (CI) against our counterpart CrossAD (CD). Best in \textbf{bold}, with ties bolded on both sides. Averaged over $3$ seeds.}
\label{tab:crossad-results}
\small
\setlength{\tabcolsep}{4pt}
\renewcommand{\arraystretch}{1.05}
\begin{tabular}{l rr rr}
\toprule
& \multicolumn{2}{c}{VUS-ROC} & \multicolumn{2}{c}{VUS-PR} \\
\cmidrule(lr){2-3} \cmidrule(lr){4-5}
Dataset & CI & CD & CI & CD \\
\midrule
PSM     & \textbf{0.7401} & 0.7364          & \textbf{0.5404} & 0.5388                   \\
MSL     & \textbf{0.8091} & \textbf{0.8091} & \textbf{0.3140} & \textbf{0.3140}          \\
SMD     & \textbf{0.8586} & 0.8581          & \textbf{0.2308} & 0.2285                   \\
SMAP    & \textbf{0.5784} & \textbf{0.5784} & \textbf{0.1444} & \textbf{0.1444}          \\
GECCO   & \textbf{0.9948} & 0.7667          & \textbf{0.6215} & 0.0814                   \\
SWAN    & \textbf{0.9554} & \textbf{0.9554} & \textbf{0.9278} & \textbf{0.9278}          \\
SWaT    & \textbf{0.6006} & 0.5938          & \textbf{0.3881} & 0.3868                   \\
\bottomrule
\end{tabular}
\end{table}

Despite its ability to recover cross-channel ruptures on the synthetic protocol of \autoref{subsec:detection-synthetic}, the CD variant brings no gain over the CI variant on any of the seven real benchmarks. MSL and SMAP yield perfect ties, since CD is exactly identical to CI on univariate time series. SWAN also produces identical scores under both variants, without an obvious explanation. On PSM, SMD, SWaT, and GECCO, CD never improves over CI. However, GECCO is a sharper failure: CI reaches VUS-PR $= 0.62$ against $0.08$ for CD, with a similar gap on VUS-ROC ($0.99$ versus $0.77$). A grid sweep over $(L, d_\mathrm{model}) \in \{16, 32, 64, 128\}^2$ reveals that CI's VUS-PR rises from $0.07$ to $0.62$ across this range, while CD stays between $0.07$ and $0.08$. Added capacity benefits CI but not CD, mirroring the capacity-robustness signature reported by~\citet{hanCapacityRobustnessTradeoff2024} in forecasting, where increasing the look-back window improves CI but degrades CD. This pattern is exactly what we predicted in \autoref{sec:empirical-evaluation}: if no labeled segment carries a cross-channel signature beyond what is already univariately detectable, the extra parameters CD uses to model cross-channel interaction have nothing to exploit, and can even lead to performance collapse.

Similar observations can be seen in the results reported by ChInf~\citep{wangChannelMattersEstimating2026} on five of our benchmarks. Their contribution improves SOTA results, despite a structurally channel-independent design: its detection score (Eq.~4 in their paper) is the maximum, across channels, of the squared gradient norm of the per-channel reconstruction loss with respect to model parameters, and therefore never couples two channels. This reiterates our observations that CI strategies outperform CD strategies on these multivariate benchmarks.

A natural extension of this experiment would be to inspect, segment by segment, which anomalies are recovered by CD but missed by CI (and conversely) rather than comparing aggregate scores alone. We leave such explorations for future work.

\section{Conclusion and Future Work}
\label{sec:conclusion}
We introduced a per-segment diagnostic that classifies each labeled anomaly as univariate, cross-channel, both, or undetected, applied it to eight standard MTSAD benchmarks, and validated it on a controlled synthetic protocol. Across the $373$ long segments, none falls in the strictly cross-channel category under any of the three correlation measures, across a sensitivity sweep over both thresholds within their reasonable range. At least half of the labeled anomaly segments deviate univariately between $89\%$ and $100\%$ of their timesteps on six of the eight benchmarks, and the synthetic protocol confirms that the diagnostic identifies cross-channel ruptures when present, with channel-dependent models (even LinearAE) recovering them while channel-independent variants cannot. On real benchmarks, the channel-dependent variant of CrossAD brings no gain over its channel-independent counterpart. The methodological consequence is that no labeled anomaly in the MTSAD benchmarks used in this study requires cross-channel modeling for detection across the reasonable range of our sensitivity sweep. Apparent gains from channel-aware architectures on these benchmarks therefore admit a univariate explanation, and these benchmarks cannot reflect whether channel-aware modeling is actually useful or not.

\citet{wuCurrentTimeSeries2021} called for abandoning NASA and SMD in 2021, and \citet{wagnerTimeSeADBenchmarkingDeep2023} independently rejected SWaT, WADI, MSL, and SMAP for MTSAD evaluation, flagging five of the eight datasets we evaluate. Yet they continue to dominate the MTSAD literature, suggesting that recommendations alone do not shift practice. Moreover, the qualitative claim of \citet{gargEvaluationAnomalyDetection2022a} that these benchmarks contain primarily temporal anomalies, echoed more recently by \citet{wenig2024}, has remained without a quantitative analysis for four years. We reiterate their recommendations for the multivariate setting: new datasets with documented anomaly mechanisms are required if channel-aware modeling needs to be evaluated, and per-segment visualizations should accompany any reported metric. More fundamentally, aggregate metrics (F1, AUC, VUS) compress per-segment behavior into a single number and obscure whether a new method actually detects anomalies that previous methods miss. We therefore endorse previous recommendations to visualize data and algorithmic outputs, and extend them: any claim that a new architecture improves over a prior one should be supported by a per-segment analysis of \emph{which} segments are newly recovered, not by a delta on an aggregate score alone.

The evaluation is restricted to labeled anomalies: if a benchmark contains genuine cross-channel events that were never labeled, our diagnostic cannot recover them. We do not claim that cross-channel modeling is unnecessary in real-world multivariate systems, but rather that many current benchmarks cannot be used to evaluate performance for cross-channel anomaly detection.

Finally, our diagnostic can serve as a structural validation test for any future MTSAD benchmark claiming to test channel-aware modeling, by checking whether its labeled segments include CROSS-CHANNEL anomalies or at least exhibit a low univariate ratio.

Two prerequisites would amplify this shift. First, dataset providers should release channel semantics, currently anonymized on most benchmarks, which alone prevents any interpretation of cross-channel dependence. Second, timestep-level labels $\bm{y} \in \{0,1\}^T$ should be replaced by per-channel labels $\bm{Y} \in \{0,1\}^{T \times C}$ documenting, at each anomalous timestep, which channels carry the anomaly. These changes could pave the way for anomaly characterization and explanation~\citep{pupae}.

\printbibliography

\end{document}